\documentclass{article}

\usepackage[final]{neurips_2020_ml4ad}

\usepackage[utf8]{inputenc} 
\usepackage[T1]{fontenc}    
\usepackage{url}            
\usepackage{booktabs}       
\usepackage{amsfonts}       
\usepackage{nicefrac}       
\usepackage{microtype}      
\usepackage{graphicx}
\usepackage{subfig}
\usepackage{mathtools}
\usepackage{amsmath,amssymb}
\usepackage{color}
\usepackage{natbib}

\newcommand{\codelocation}[1]{\small\url{https://github.com/Manojbhat09/Trajformer}.}

\title{Trajformer: Trajectory Prediction with Local Self-Attentive Contexts for Autonomous Driving}

\author{%
	Manoj Bhat$^{1}$ \qquad Jonathan Francis$^{1,2}$\thanks{Correspondence.} \qquad Jean Oh$^{1}$\\ $^1$Carnegie Mellon University \qquad $^2$Bosch Research Pittsburgh\\
	\texttt{\{mbhat, jmf1, hyaejino\}@andrew.cmu.edu}
}

\begin{document}

\maketitle

\begin{abstract}
Effective feature-extraction is critical to models' contextual understanding, particularly for applications to robotics and autonomous driving, such as multimodal trajectory prediction. However, state-of-the-art generative methods face limitations in representing the scene context, leading to predictions of inadmissible futures. We alleviate these limitations through the use of self-attention, which enables better control over representing the agent's social context; we propose a local feature-extraction pipeline that produces more salient information downstream, with improved parameter efficiency. We show improvements on standard metrics (minADE, minFDE, DAO, DAC) over various baselines on the Argoverse dataset. We release our code at:~\codelocation.

\end{abstract}


\section{Introduction}
\label{section:introduction}

Precise contextual understanding and feature-extraction is critical in developing reliable models for, e.g., trajectory prediction in autonomous driving scenarios. Typically, generative model classes can be seen as a composition of encoder and decoder structures, where the encoder is responsible for mapping observations to some intermediate feature representation and the decoder is responsible for leveraging this representation for generating a set of future trajectory predictions. Because the trajectory predictions are conditioned on this intermediate representation, we can also say that the quality of the predicted trajectories is a function of the comprehensiveness of the feature representation itself, in terms of its ability to encode all of the salient events in the world in which the agent operates. 

\cite{park2020diverse} highlight a distinction between two major types of environmental context, described with respect to the ego-agent: scene-to-agent context and agent-to-agent context. The former characterises the static properties of the environment, such as admissible driving regions and infrastructural road elements (e.g., traffic lights), and the latter characterises more dynamic elements of the environment, such as pedestrians, other vehicles, and other moving obstacles. Whereas various works have addressed the use of scene-to-agent context as in differentiable driveable-area maps, efficient encoding of agent-to-agent context remains an open challenge for autonomous driving, despite being widely studied for human trajectory prediction \citep{alahi2016social, gupta2018social, sdd, vemula2018social}. 

Relating to agent-to-agent context, specifically, the ego-agent must model its state, amid the set of time-dependent social relationships between all the other agents in the scene. The ability to model these social interactions is crucial, because it informs the ego-agent about the behaviours that are appropriate, given the scene-to-agent context; furthermore, these interactions also inform the ego-agent about the proper social etiquette for driving alongside other agents on the road. Indeed, both aspects of the agent-to-agent context have direct impacts on the predicted trajectories, downstream. However, the highly multimodal nature of the scene complicates models' ability to extract salient information about agent behaviour and social etiquette, leading to the predictions of inadmissible futures in autonomous driving \citep{casas2020importance, park2020diverse}. In this work, we propose the use of self-attention~\citep{vaswani2017attention} that serves as a better structure of modeling the important latent factors in dynamical scenes. 

We propose \texttt{Trajformer}, an end-to-end model that addresses prior limitations in modeling the social contextual relationships, in multimodal trajectory prediction for autonomous driving. We achieve this through the use of a self-attention-based encoding structure, allowing for better characterisation of agent behaviour and social etiquette, by providing focused local features given other objects in the scene (e.g., dynamic obstacles, pedestrians, other vehicles). We validate our approach on the Argoverse dataset \citep{chang2019argoverse} and show substantial performance improvement over baselines and ablations, across standard metrics (minADE, minFDE, DAO, and DAC). Additionally, we show significantly improved model parameter-efficiency, relative to the state-of-the-art. We publish our tools and code for replicating our experiments:~\codelocation





\section{Related Work}
\label{section:related_work}

\paragraph{Encoder structures for multimodal trajectory prediction.}

Social contextual modeling has been widely studied in the area of human trajectory prediction \citep{alahi2016social, gupta2018social, sdd, vemula2018social, giuliari2020transformer}. \cite{giuliari2020transformer} introduced a method for utilizing transformer models~\citep{vaswani2017attention} to produce pedestrian trajectory predictions with multiple mode support. We propose a similar encoder with spatial priors, in the form of projected embeddings from map croppings~\citep{dosovitskiy2020}, with specific application to modeling the social context in autonomous driving scenarios. With our performance improvements, we illustrate the contribution of transformer models in the crucial feature-extraction step -- linking annotation-free diverse prediction with self-attention, for long-range dependency modeling~\citep{vaswani2017attention, 2016liattentive}. 

Another promising encoding structure for forecasting models is in the use of graph-based encoders. \cite{liang2020learning} represented the scene context (i.e., lane center-lines) as a discrete graph and, given past poses therein, they used attention to extract salient features for predicting trajectory futures. 
Similarly,~\cite{messaoud2020trajectory} and \cite{monti2020dag} highlight the components necessary for predicting futures, such as: graph-based feature extractors, image feature extractors, network classes to perform fusion of these information modalities, and the prediction heads themselves. We propose a more integrated approach -- with a single self-attention-based backbone, for learning social behavior+etiquette and generating diverse+admissible futures. Further, while these approaches focused on the spatial context, we pursue performance gains through modelling the social context.

\paragraph{Importance of model size for trajectory prediction.}

\cite{carion2020endtoend} discuss the importance of understanding the trade-off between model capacity and performance, in the context of object detection and semantic segmentation tasks; here, image features are extracted from a transformer-based backbone network and joined with a positional encoding representation. \cite{dosovitskiy2020} proposed how image croppings can be directly utilized for feature-extraction with self-attention. In this work, we are inspired by these discussion and make specific application to multimodal trajectory prediction for autonomous driving.


\begin{figure}[!t]
\centering
\subfloat[]{ 
\includegraphics[width=.6\linewidth]{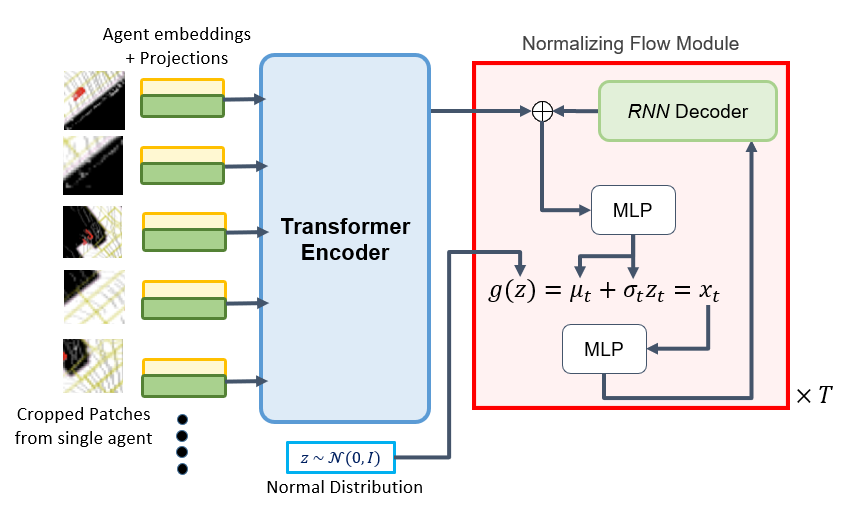}
\label{fig:CrossAgentModule}
}
\subfloat[]{
\includegraphics[width=.32\linewidth]{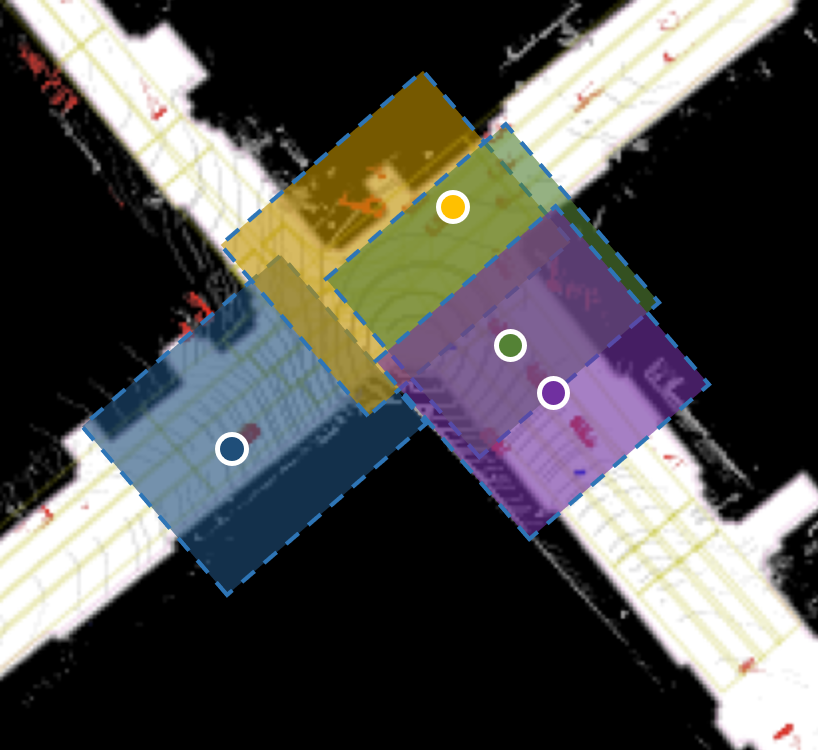}
\label{fig:AgentInteractionModel}
}

\vspace{-3mm}
\caption{
(a) Model architecture with Agent pose embeddings, cropped image and positional embeddings fused for input to the transformer encoder and Flow based decoding for producing $T$ future poses. This reduced architecture can be useful for Trajectory prediction for embedded platforms in Robotic applications.  
(b) Depiction of Patch croppings produced from the BEV image. The different colors indicate different Agents in the scene. And the colored area is a fixed $K \times K$ pixel size for each such crops. These patches are then again cropped into $16 \times 16$ patches and linearly projected to produce projections at the input of the transformer model. 
}
\label{fig:model_modules}
\end{figure}






\section{Local Self-attention for Trajectory Prediction}
\label{section:approach}

In this work, we unify a self-attention-based~\citep{vaswani2017attention} encoder structure with a normalising flow-based decoder structure~\citep{park2020diverse}, for multimodal trajectory prediction in autonomous driving. Our method is illustrated in figure \ref{fig:CrossAgentModule} and is further described below. 

\paragraph{Model structure.} To encode the social factor between $A$ agents, we combine their individual past-trajectory encodings, through consecutive additive and multiplicative fusion~\citep{liu2018learn}, to generate $A$ embeddings. For each of these agent embeddings, $t$ past time-step poses are raised to $N$-dimensional vectors and are combined with patch and positional embeddings, for each time-step. In this way, we generate an $N\times~t$-dimensional input for the transformer model. 

Let $\boldsymbol{S} \equiv\left\{S^{1}, S^{2}, \ldots, S^{A}\right\}$ denote the set of agent trajectories for $A$ agents in a given scene, with $S^a$ being the concatenation of past trajectory segment $S^{a}_\text{past}$ and ground-truth future trajectory segment $S^{a}_\text{future}$. Here, a single step is indicated by $S^a_t \in \mathbb{R}^2$, for agent $a$ and time-step $t$. Thus, $\boldsymbol{S}_\text{past}$ is the collection of past trajectory segments for all agents, and the observation set of all 3-second past trajectories is denoted by $\mathcal{O} \equiv\left\{\boldsymbol{S}_{\mathrm{past}}, \boldsymbol{\Phi}, \boldsymbol{\phi}\right\}$, where $\Phi$ is a scene embedding~\citep{park2020diverse} and $\phi$ is the positional embedding~\citep{vaswani2017attention}. We want to model the posterior distribution over future trajectories, for all agents in the scene snapshot, $q(\boldsymbol{S}_{\text{pred}} | \mathcal{O})$. 


Past agent trajectories are projected to a (higher) $d$-dimensional space, in preparation for input to the transformer, i.e., $\mathbf{e}_{obs}^{a}=\texttt{MLP}_{proj}(\mathrm{S}^{{a}})$. For encoding the contextual information, we extract an $m^2$ pixel neighbourhood image patch, centered around each vehicle, from the \textit{birds-eye-view} (BEV) map of the scene: $\mathbf{e}_{patch}^{a}$. The BEV map contains coloured objects and superimposed LiDAR points. Figure \ref{fig:AgentInteractionModel} illustrates the agent-wise pixel neighbourhood, which capture local contextual information. 

We perform sine-distance positional encoding of the map representation, which is then added to each agent's flattened sequence of patch vectors. We then calculate a fused representation, combining the local environment information and state history, as a \textit{Hadamard} product between each agent's past trajectory embedding and its corresponding scene context~\citep{liu2018learn}: $$\mathbf{e}_{fused}^{a}=\mathbf{e}_{obs}^{a}~\odot~[\texttt{ENC}_{pos}(\mathbf{e}_{map}^{a})~+~\texttt{FLATTEN}(\mathbf{e}_{patch}^{a})]$$





These fused representations are fed to a standard transformer encoder, which contains alternating layers of multi-headed self-attention and MLP blocks. The output is a set of latent codes -- one for each agent: $\mathbf{c}^{a}_{latent}=\texttt{ENC}_{tr}(\mathbf{e}_{fused}^{a})$, with $\mathbf{c}^{a}_{latent} \in \mathbb{R}^{D\times A}$ with hyperparameter $D$.


The normalizing-flow-based generative decoder features an implicit auto-regressive design and performs a differentiable and bijective mapping, from the latent codes to the set of agent-wise trajectory predictions \citep{park2020diverse, rhinehart2019precog} (see figure \ref{fig:CrossAgentModule} for illustration): $g_{\theta}(z_{t} ; \mu_{t}, \sigma_{t})=\sigma_{t} \cdot z_{t}+\mu_{t}=S_{pred,t}^{a}$, with $z_{t} \sim \mathcal{N}(\mathbf{0}, I) \in \mathbb{R}^{2}$. Here, $\theta$ is the set of model parameters and $\mu_{t} \in \mathbb{R}^{2}$ and ${\sigma}_{t} \in \mathbb{R}^{2 \times 2}$ are projected parameters. Iterating through time, we get the predictive trajectory $S_{\text {pred }}^{a}$ for each agent. By sampling multiple instances of $z_{\text {pred }}$ and mapping them to trajectories, we get various hypotheses of future.



Following \cite{park2020diverse}, we train our model according to the symmetric cross-entropy objective~\citep{rhinehart2018r2p2}, with the annotation-free discrete grid map as the (estimated) prior distribution $\hat{p}$ and with a model degradation coefficient of $\alpha = 0.5$. 


\section{Experiments}
\label{section:experiments}

We benchmark two instances of our approach, \texttt{Trajformer-12} and \texttt{Trajformer-24}, with respectively 12 and 24 layers in the transformer encoder. We set the size of the trajectory encoder projection $d$ to be 1024 ($\texttt{MLP}_{proj}$ is a single-layer projection), pixel neighbourhood width/height $m$ to be 16, and the dimension of the latent code $D$ to be 256. We choose a batch size of 128 and train with Adam optimizer. We use linear learning rate warm-up and decay. It takes 3 days to train each model on a NVIDIA 1080 Ti GPU device, with batch data processed as in~\cite{park2020diverse}, from the \textit{Tracking} split of the Argoverse dataset~\citep{chang2019argoverse}.









\section{Results}
\label{section:results}

\begin{figure}[!t]
\centering
\includegraphics[width=.8\linewidth]{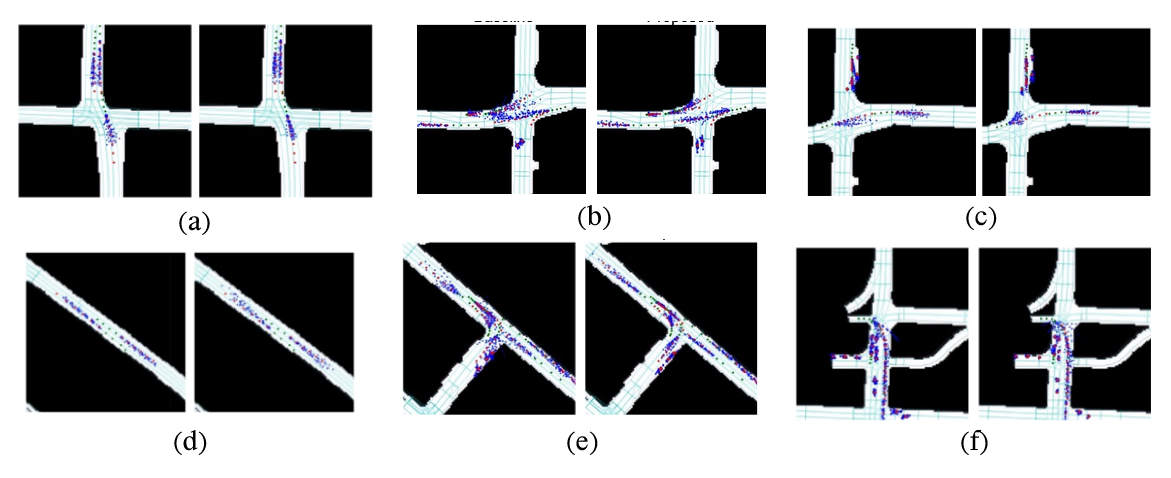}

\caption{Qualitative illustration of model performance [right], compared to best-performing baseline \citep{park2020diverse}[left]. Observations: (a) more precise \& confident on straight in-lane trajectories; (b) more confidence in the maneuver, indicated by a cluster of trajectories; (c) more confident and diverse alternative maneuvers, with attention to standing vehicles, due to simple intersection map lane-start/end prior; (d) equivalent lane-change maneuver on empty roads, due to attention on immediate local activities; (e) more conservative turning maneuver with more agent-activity;
(f) reduced confidence in strong curve lane-change maneuvers.}
\label{fig:qualitative}
\end{figure}

\begin{table}[!t]
\caption{Comparison of improvements over baseline models on Argoverse. The metrics are abbreviated as follows: \textsc{minADE}(\textbf{A}), \textsc{minFDE}(\textbf{B}),  \textsc{rF}(\textbf{C}), \textsc{DAO}(\textbf{D}), \textsc{DAC}(\textbf{E}). Improvements indicated by arrows. $*$: larger is better, as long as \textbf{A} and \textbf{B} are small.}
\setlength{\tabcolsep}{0.1pt}
\scriptsize
\centering
\begin{tabular}{p{5cm}p{0.9cm}p{0.9cm}p{0.9cm}p{0.9cm}p{0.9cm}}
\toprule
  & A~($\downarrow$) & B~($\downarrow$) & C~($\uparrow$)* & D~($\uparrow$)* & E~($\uparrow$)*\\
 \midrule
 {\textsc{LSTM}} & 1.441 & 2.780 &  1.000 &  3.435 & 0.959  \\
  {\textsc{CSP~\citep{park2020diverse}}} & 1.385 & 2.567 &  1.000 & 3.453 & 0.963\\
  {\textsc{MATF-D~\citep{zhao2019multi}}}& 1.344 & 2.484 &  1.000 &  1.372 & 0.965 \\
    {\textsc{DESIRE~\citep{lee2017desire}}} & 0.896 & 1.453 & 3.188 & 15.17 & 0.457 \\
     {\textsc{MATF-GAN~\citep{park2020diverse}}} & 1.261 & 2.313 &  1.175 & 11.47 & 0.960 \\
       {\textsc{R2P2-MA~\citep{rhinehart2018r2p2}}}  & 1.108 & 1.270 & 2.190 &  37.18 & 0.955  \\
         {\textsc{DATF~\citep{park2020diverse}}} & 0.730 & 1.124 &  3.282 &  28.64 &  0.968 \\
         \hline
             {\textsc{Trajformer-12}~(ours)} & {0.684} & {0.885} &  {3.359} &  {27.71} &  0.972 \\
                 {\textsc{Trajformer-24}~(ours)} & \textbf{0.621} & \textbf{0.719} &  \textbf{3.868} &  \textbf{28.21} &  \textbf{0.973} \\
\midrule
\end{tabular}
\label{tab:main_experiment}
\vspace{-6mm}
\centering
\end{table}

\begin{table}[!t]
\caption{Model size comparison (with optimizer state), in megabytes (MB) and number of parameters. 
}
\begin{center}
\scriptsize
\begin{tabular}{lcccc}
\toprule
Model-layers & & \textsc{Size (.tar)} & & \textsc{\#Params} \\
\hline
\shortstack{\textsc{DATF~\citep{park2020diverse}}} & & \shortstack{{4.7}~MB} & & \shortstack{{462}K} \\
\hline
\shortstack{\textsc{Trajformer-12} (ours)} & & \shortstack{2.1~MB} & & \shortstack{164K} \\
\shortstack{\textsc{Trajformer-24} (ours)}  & & \shortstack{2.9~MB} & & \shortstack{192K} \\
\bottomrule
\end{tabular}
\end{center}
\vspace{-8mm}
\label{tab:cross_agent_exp}
\end{table}


Quantitative results are summarized in Table \ref{tab:main_experiment}, and some qualitative results are shown in \ref{fig:qualitative}. We observe a new state-of-the-art performance in our model, compared to \cite{park2020diverse}, in both qualitative and quantitative results. Most of the maneuvers have been refined by the model. An interesting observation in~figure~\ref{fig:qualitative}b suggests that rule-based maneuvers, such as the right-of-way in intersections, are learned and followed by model (the left vehicle gains the right-of-way).
We also note a significant failure mode: where, if the velocity of a particular agent in a particular frame is high, the target 6 trajectory points are spaced uniformly and with a much greater \textit{intra}-point distance (2x), compared to the average spacing between each data-point in a trajectory from the Argoverse dataset. Here, the model fails to predict the appropriate spacing, despite producing the right modes of trajectories. The reason for this phenomenon is hypothesised to be the size of the local neighbourhood context that was chosen while training the model. Compared to DATF \citep{park2020diverse}, the model is significantly lighter in time-complexity and memory-intensity, due to the social attention and scene attention blocks provided by the transformer encoder. The \texttt{Transformer-12} and \texttt{Transformer-24} model instances did not vary significantly in the qualitative and quantitative results: the aforementioned observations remain true for both models.


\section{Conclusion}
\label{section:conclusion}

In this paper, we proposed \texttt{Trajformer}, an end-to-end model that addresses prior limitations in modeling the social contextual relationships, in multimodal trajectory prediction for autonomous driving. We introduced the use of self-attention-based encoding for applications to autonomous driving, allowing for better characterisation of agent behaviour and social etiquette, given other objects in the scene (e.g., dynamic obstacles, pedestrians, other vehicles). Our approach uses a single self-attention-based backbone, for learning social behavior+etiquette and generating diverse+admissible futures. We validated our approach on the Argoverse dataset \citep{chang2019argoverse} and showed substantial performance improvement over baselines and ablations, across standard metrics (minADE, minFDE, DAO, and DAC). Additionally, we show significantly improved model parameter-efficiency, relative to the state-of-the-art. 



\newpage

\small
\bibliographystyle{plainnat}
\bibliography{references.bib}  




\end{document}